\documentclass[10pt, a4paper]{article}

\usepackage[final]{lrec-coling2024} 

\usepackage{CJKutf8} 

\title{Event-enhanced Retrieval in Real-time Search}

\name{Yanan Zhang, Xiaoling Bai, Tianhua Zhou} 

\address{Tencent Search, Platform and Content Group \\
         \{yananzhang, devinbai, kivizhou\}@tencent.com\\}

\abstract{
The embedding-based retrieval~(EBR) approach is widely used in mainstream search engine retrieval systems and is crucial in recent retrieval-augmented methods for eliminating LLM illusions. However, existing EBR models often face the ``semantic drift'' problem and insufficient focus on key information, leading to a low adoption rate of retrieval results in subsequent steps. This issue is especially noticeable in real-time search scenarios, where the various expressions of popular events on the Internet make real-time retrieval heavily reliant on crucial event information. To tackle this problem, this paper proposes a novel approach called EER, which enhances real-time retrieval performance by improving the dual-encoder model of traditional EBR. We incorporate contrastive learning to accompany pairwise learning for encoder optimization. Furthermore, to strengthen the focus on critical event information in events, we include a decoder module after the document encoder, introduce a generative event triplet extraction scheme based on prompt-tuning, and correlate the events with query encoder optimization through comparative learning. This decoder module can be removed during inference. Extensive experiments demonstrate that EER can significantly improve the real-time search retrieval performance. We believe that this approach will provide new perspectives in the field of information retrieval. The codes
and dataset are available at \url{https://github.com/open-event-hub/Event-enhanced_Retrieval}.
 \\ \newline  }

\begin{document}
\begin{CJK*}{UTF8}{gbsn} 

\maketitleabstract

\section{Introduction}

The embedding-based retrieval~(EBR) approach has gained attention since its introduction and is widely used in the recall systems of mainstream search engines. It also plays a crucial role in recent methods aimed at mitigating the hallucinations of large language models through retrieval-augmented techniques, with many LLM application frameworks such as Langchain\footnote{https://github.com/langchain-ai/langchain} providing such tutorials. Compared to traditional term-level retrieval algorithms like BM25~\citep{bm25-2009}, EBR can effectively capture semantic similarity beyond word frequency or term matching, thus better handling synonyms, near-synonyms, and context-related semantic relationships. However, efficiently retrieving the most relevant documents from billions of documents remains a daunting challenge.

\begin{figure}
    \centering
    \includegraphics[width=0.99\linewidth,keepaspectratio]{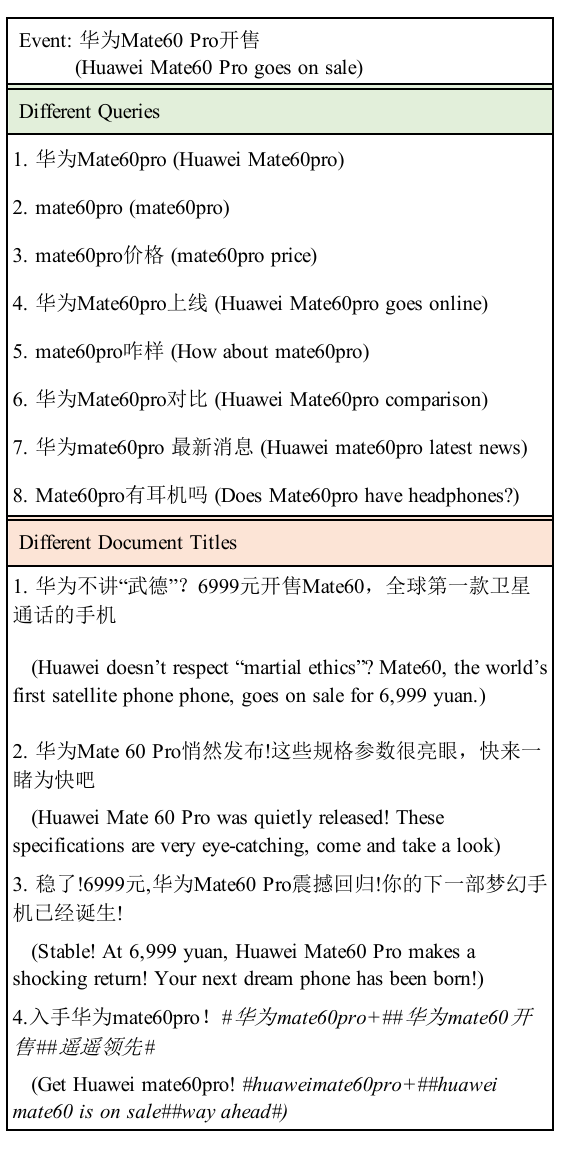}
    \caption{An event corresponds to various queries and documents. Most queries are always concise, focusing on the key information of the event, and often contain abbreviations, omissions, grammatical irregularities, etc. For example, in the second query ``mate60pro'', ``Huawei'' is omitted, and ``Mate'' is entered as ``mate''. The document title is lengthy, contains redundant information, and the expression style is diversified. In the third title, the action ``稳了~(Steady)'' lacks a subject and is an unconventional syntax. The fourth title contains a lot of tags with ``\#''. It is therefore difficult to relate queries to documents. The data here is from the real world.}
    \label{fig:demo}
\end{figure}

One of the main challenges faced by existing EBR models is the ``semantic drift'' problem,  i.e., the semantics of the model encoding deviates from the given context, lacking attention to key information. This problem becomes particularly pronounced in real-time search~\citep{real_time_search} scenarios, where users tend to input shorter queries, typically keywords or phrases about an event, to quickly obtain information about the event. 
On the other hand, there are multiple ways of expressing the same event on the internet, considering different media sources and even self-media. Moreover, documents are generally longer than queries, and even if only considering the title, they still contain a lot of less important information, as shown in Figure~\ref{fig:demo}. The highly asymmetric information between queries and titles makes real-time retrieval of event documents more difficult. Existing research has not paid special attention to the differences and difficulties of real-time search compared to other searches. On the one hand, attempts have been made to improve the embedded representation performance by introducing more massive data and models with larger parameters~\citep{ni-etal-2022-large, INSTRUCTOR,wang2022text,li2023general,bge_embedding}, to achieve a ``miracle'' effect, which in fact does lead to an improvement, but the pursuit of lower cost and smaller model parameters deserves to be considered all the time. On the other hand, a large number of data augmented approaches~\citep{wei-zou-2019-eda,liu-etal-2021-fast,wu-etal-2022-esimcse,chuang-etal-2022-diffcse,tang-etal-2022-augcse} are used, such as token duplication, substitution, etc., but these schemes pay little attention to events~(the central secret of real-time search) and are not sufficient to cope with the complexity of the scenario.

To address this pressing problem, we propose an event-enhanced retrieval~(EER) method, which builds on the traditional EBR dual encoder model that utilizes $<query, title>$ pairs. We introduce various hard negative mining techniques and apply supervised contrastive learning~\citep{gao2019representation} to improve the performance of the encoders. To further widen the gap between positive and negative example samples, we also incorporate pairwise learning, enabling the encoder to better focus on relative order and enhance robustness. To address the $<query, title>$ pairs information asymmetry and the abundance of noisy information in titles, we creatively introduce a decoder structure outside the title-side encoder. The decoder aims to receive the encoded title information and extract event triplets through a generative task, facilitating the title-side encoder to focus more on the event information. We also employ keyword-based prompt learning to make the generated content more controllable. The events generated by the decoder also deserve attention from the query, as they represent crucial event information from the title. Therefore, the $<query, generated-event>$ pairs interact through supervised contrastive learning to boost the performance of the query encoder. It is worth noting that this decoder is only used in the training step to enhance the understanding of the event, while it can be removed in the inference phase, and EER will revert to the traditional dual-tower model with no impact on latency.

The main contributions of this work are summarized as follows:
\begin{itemize}
\item To our knowledge, EER is the first approach that tackles the ``semantic drift'' issue in real-time search scenarios, aiming to enhance the retrieval of event documents. 
\item Building upon the traditional dual-tower model, we introduce a generation task specialized for events in titles. By employing loss functions that emphasize the attention of both encoders to the events, we achieve state-of-the-art performance for the encoders.
\item Numerous experiments and analyses have indicated the strong merit of EER.

\end{itemize}

\begin{figure*}[t]
    \centering
    \includegraphics[width=\linewidth]{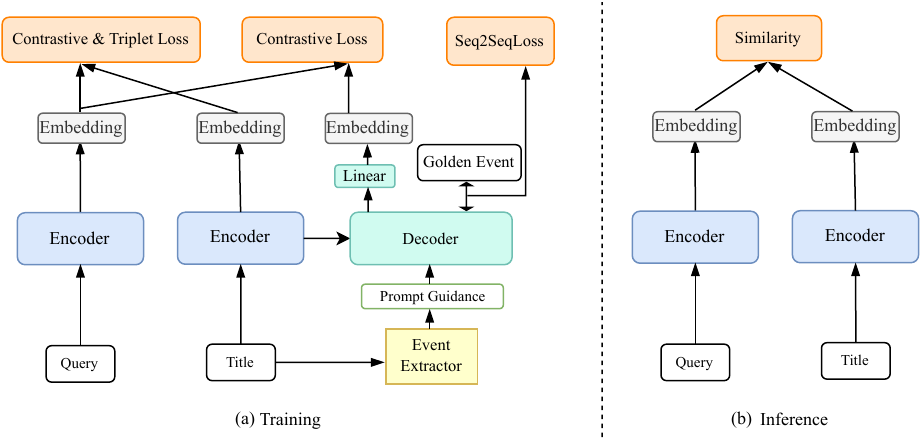}
    \caption{Architecture of the proposed EER model.}
    \label{fig:ar}
\end{figure*}

\section{Methods}

\subsection{Overview}

In this section, we describe the proposed EER model in detail. As shown in Figure~\ref{fig:ar}(a), the most basic structure of the model is a dual tower that encodes the query and the document title, respectively. Further, we focus on the extraction and usage of document event information, i.e., we add a prompt learning-based decoder module after the encoder of the document and influence the representation performance of both encode modules through the loss feedback, to improve the retrieval ability of the event documents under real-time search. Finally, we mention the difference between the inference and train pipeline. As shown in Figure~\ref{fig:ar}(b), the newly added decode module can be taken off in the inference stage, and the model is restored to the traditional dual towers.

\subsection{Encoder}

\subsubsection{Hard Negative Sampling}
Hard negative sampling~\citep{robinson2021contrastive} has become an important method over the years. Compared to random negative sampling, hard negative sampling is able to pull off the gap between positive and negative samples in a more targeted way. We do hard negative sampling using a knowledge augmentation-based approach and a semantic mining-based approach.

\textit{\textbf{Knowledge Augmentation}}: To improve the robustness of EER sentence representation, we perform data augmentation of queries and headings following EDA~\citep{wei-zou-2019-eda}  before feeding pairs of sentences into the encoder. We mainly used three data augmentation strategies, including entity replacement~(encyclopedia), random token deletion and replication, and token reordering. We speculate that~(1) Using entity replacement is an effective strategy to create semantically similar phrases with different tags, which helps the model capture keyword similarity rather than syntactic similarity. (2) The random deletion strategy can mitigate the impact of frequent words or phrases. (3) The shuffling strategy can reduce the sensitivity of the sentence encoder to position changes.

\textit{\textbf{Semantic  Mining}}: We fine-tune an encoder model based on existing training data to encode all collected titles and store them into a Faiss~\citep{johnson2019billion} vector index library. Then, for each query, we retrieve its $k$ neighboring titles using semantic similarity. 
For these $k$ titles, we randomly keep up to $m~(m\ll k)$ of them whose relevance~(cosine similarity) is between a predefined upper and lower bound. Through this operation, we get tough but low-relevance negative titles.

\subsubsection{Constrastive Learning}\label{sec:qtcl}

To help our model learn sentence representations better and alleviate the problem of vector space collapse, we utilize contrastive learning~\citep{gao2019representation} techniques to pull the vector distance closer between a query and its specified positive titles while pushing the distance further between the query and negative titles. The negative titles including randomly sampled and the hard negatives mentioned earlier, mean that the hard negatives of one query will also be shared with other queries, which may further augment the scale of negatives. The comparative learning loss of query and title can be formulated as:

\begin{equation}
\mathcal{L}_{cl_{qt}} = -\log\frac{e^{cos(\mathbf{h}_i,\mathbf{h}_i^+})/\tau}{\sum^N_{j=1}\left ( e^{cos(\mathbf{h}_i,\mathbf{h}_j^+)/\tau} + e^{cos(\mathbf{h}_i,\mathbf{h}_j^-)/\tau}  \right ) }
\end{equation}

where $\mathbf{h}_i$, $\mathbf{h}_i^+$ and $\mathbf{h}_i^-$ are the representation of the $i$-th query, its positive sample and its negative sample respectively, $cos(\cdot)$ is the cosine similarity, $\tau$ is a temperature hyper-parameter and $N$ is batch size.

\subsubsection{Pairwise Learning}

Consider a query and its positive and negative titles are constructed according to the pairwise formula and the ordinal relationship between positive and negative examples is also important, we utilize triplet loss~\citep{wang2014learning} to strengthen this kind of relevance ranking. Inspired by Sentence-BERT~\citep{reimers-2019-sentence-bert}, the loss can be illustrated as:

\begin{equation}
\mathcal{L}_{pair_{qt}} = \max(0, \epsilon + cos(\mathbf{h}_i,\mathbf{h}_i^-) - cos(\mathbf{h}_i,\mathbf{h}_i^+))
\end{equation}

where $\epsilon$ is the margin to ensure the similarity of $(\mathbf{h}_i,\mathbf{h}_i^+)$ is at least closer than $(\mathbf{h}_i,\mathbf{h}_i^-)$, which is set as $0.1$ to avoid over-fitting.

\subsection{Generative Decoder}

In this section, we introduce methods to help our model enhance event information awareness by introducing decoder modules and subtasks at the title.
\subsubsection{Event Extraction}

The event information is one of the core components of a headline. Extracting which phrase directly affects what information we encode in the sentence representation.
Given the characteristics and difficulties of Chinese news headlines, we mainly use two existing methods to extract event information: semantic role labeling and dependency syntactic parsing of a sentence. By utilizing LTP toolkit~\citep{che-etal-2021-n}, we can easily obtain the semantic role labeling of a sentence, and extract the subject-predicate-object structure, subject-predicate structure, and predicate-object structure in order. If the semantic role tags are empty, we use dependency syntactic parsing methods to extract event triples centered around the predicate, including subject-predicate-object, post-verb object with attributive, and subject-predicate-verb-object with preposition. Additionally, we use the Title2Event~\citep{deng20222event} dataset to fine-tune a supervised Seq2SeqMRC model, a pipeline model that replaces the argument extractor with a sequence-to-sequence MRC model using mT5-base~\citep{mt5}. In practice, we find that the Seq2SeqMRC model performs better than LTP, and we choose it as the event extraction tool.

\subsubsection{Generation Learning with Prompt Guidance}

We added a decoder module to the original two-tower encoder model to undertake the event information generation subtask. As shown in Figure~\ref{fig:ar}(a), the decoder is another $12$-layer RoBERTa~\citep{YinhanLiu2019RoBERTaAR}, whose parameters we initialise in the same way as the encoder, inspired by BERT2BERT~\citep{rothe2020leveraging}, so that the parameters of the decoder are isomorphic to those of the encoder, reducing maintenance costs.
During the training stage, similar to any standard sequence-to-sequence transformer architecture, the original titles are fed into the encoder, and events, marked as $E$ and extracted from the titles, serve as the ground truth and are fed into the decoder.

Further, to ensure that important information is not overlooked, we leverage prompt learning techniques in the generation task. Unlike previous work, we use adaptive keyword templates to guide event generation. One of the keyword templates $T$ is similar to ``In \texttt{[X]}, the object is \texttt{[MASK]}, the trigger is \texttt{[MASK]}, and the topic is \texttt{[MASK]}'', where ``\texttt{[X]}'' is the text of the title. In this form, we mimic the masked language model pre-training task so that the model can perceive the object, subject, and trigger of the event. The final input to the decoder is labeled ``\texttt{[CLS]}T E\texttt{[SEP]}''.

The event generation loss is formulated as:
\begin{equation}
\mathcal{L}_{gen}= -\sum_{i=1}^{N} \sum_{t=1}^{T} y_{i,t} \log \hat{y}_{i,t}
\end{equation}

where the variable $N$ represents the number of samples in the dataset, and $T$ denotes the length of the target sequences. The elements $y_{i,t}$ and $\hat{y}_{i,t}$ are the true and predicted probabilities of the target token at position $t$ in the $i$-th sample, respectively.

\subsubsection{Relevance Learning Between Query and Event}

Events generated by the decoder module can be regarded as a ``condensed version'' of the title because it represents the critical information of the title, has less information noise, and is shorter than the title. Therefore, compared with titles, events can alleviate information asymmetry and are worthy of being used to interact with queries to optimize model performance. Specifically, when a query is related to a title, we consider the query to be related to the event corresponding to that title. We add a sub-task to characterize this similarity, where positive samples are events generated from positive titles, and negative samples are other events in the batch, also learned through contrastive learning loss, which can be formulated as:

\begin{equation}
\mathcal{L}_{cl_{qe}} = -\log\frac{e^{cos(\mathbf{h}_i,\mathbf{h}_i^+})/\tau}{\sum^N_{j=1}\left ( e^{cos(\mathbf{h}_i,\mathbf{h}_j^+)/\tau} + e^{cos(\mathbf{h}_i,\mathbf{h}_j^-)/\tau}  \right ) }
\end{equation}

where $\mathbf{h}_i$, $\mathbf{h}_i^+$ and $\mathbf{h}_i^-$ are the representation of the $i$-th query, its positive sample and its negative sample respectively. The difference from~\ref{sec:qtcl} is that the samples here are replaced by events.

\subsection{Total Loss}

In summary, our training task consists of three parts: query-title relevance learning, event generation learning, and query-event relevance learning. Among them, query-title relevance learning consists of both contrastive learning and pair learning. The loss function can be completely formulated as:

\begin{equation}
\mathcal{L}_{total} = \mathcal{L}_{cl_{qt}} +  \mathcal{L}_{pair_{qt}} + \mathcal{L}_{gen}+\mathcal{L}_{cl_{qe}}
\end{equation}

\subsection{Inference Pipeline}

Indeed, the purpose of adding a decoder module to the title to implement event extraction and interact with the two encoders is to assist in optimizing the performance of the two encoders. On the other hand, a complex model structure brings inference latency and reduced ease of use, which also needs to be considered. Therefore during the inference stage, we only need vector representations of queries and titles, and the decoder module can be removed, as shown in Figure~\ref{fig:ar}(b). EER reverts to the traditional dual-tower model, which means there is no significant change in time consumption and ease of use.

\section{Experiments}
\subsection{Experimental Settings}
\subsubsection{Datasets}

Considering that there is no retrieval dataset tailored specifically for real-time search, we produce one ourselves and make it publicly available. 
The dataset is gathered from the massive user logs of Tencent QQ Browser Search, thus the vast majority of data is in Chinese and the authenticity of the data is guaranteed.
The query patterns in this data are varied. At the same time, the titles come from various types of documents such as text~(including User Generated Content), video~(including mini video), etc., covering 23 news categories such as current affairs, sports, finance stock, technology, society, and entertainment. As shown in Figure~\ref{fig:demo} earlier, the characteristics of query and title are very distinct from the existing benchmark, and the experimental results in Section~\ref{sec:oresult} later demonstrate the uniqueness of this data.

Specifically, relevant documents are labeled as 1 and irrelevant ones as 0 in the data. For the training data, we sample $<query, title>$ pairs from the user logs of real-time search requests~(using an existing intent recognition tool) over the past six months. We start with an initial filtering of the data using some of the features that the logs already contain, such as quality score, authority, and harmfulness and make an effort to remove personal information about private individuals. For each pair, multiple rules are constructed to automatically annotate the dataset using user behaviors such as clicks, document browsing duration, page flipping, and query reformulation, combined with relevance features such as BM25 and term matching rate. This method enables mining a large amount of data from user logs. 

To prevent ``data leakage'', we use the same sampling method as the training data for the test data, but we choose a different timeframe - specifically, the month following the training data. This helps to ensure that the test data is independent of the training data. Given the enormous size of the log data, the probability of sampling data from the same user is low. Furthermore, we concentrate on event-related queries and documents, which are updated rapidly due to the constant flow of new information. This means that users' interests can shift quickly, and we need to ensure that our data reflects these changes accurately.
Considering the limitations of rule-based annotation, we chose to ensure the data quality through expert annotation on a crowdsourcing platform. We write an $8$-page annotation document that includes numerous examples to illustrate the standard. Additionally, we develop an annotation tool with a search function to assist experts in making judgments. Each data is double annotated, and this data will be rechecked when the double annotations are inconsistent to ensure that the final accuracy of the test data reaches $95\%$. Detailed data statistics are shown in Table~\ref{tab:dataset_statistics}.

\begin{table}[t]
\centering

\begin{tabular}{c | c c c}
\hline
Dataset & Queries & Titles & Q-T pairs \\
\hline
Training & 2,964,077 & 5,323,681 & 10,319,501 \\
Testing & 1,096 & 4,733 & 10,2279 \\
\hline

\end{tabular}
   
\caption{The statistics of the dataset.}
\label{tab:dataset_statistics}
\end{table}

\subsubsection{Evaluation Metrics}
We adopt Recall@$k$~(R@$k$)~\citep{jegou2010product}, Mean Reverse Ranking~(MRR)~\citep{Craswell2009}, and AUC~\citep{FAWCETT2006861} to compare models. R@$k$ measures the ratio of queries in which the correct template is within the top-$k$, while MRR computes the mean reverse of the correct template. Both metrics tend to focus on positional relationships. AUC is used to observe the ability to discriminate between the full range of positive and negative samples. In particular, we track R@$10$ and MRR@$10$ as a general indicator of ranking performance.

\subsubsection{Baselines and Parameter}
We selected several representative baseline methods:

\textit{\textbf{BM25}}~\citep{bm25-2009} A classic algorithm used to evaluate the relevance between a query and a document. It considers factors such as term frequency, document length, and inverse document frequency to determine the relevance of a document to a given query.

\textit{\textbf{Sentence-BERT}}~\citep{reimers-2019-sentence-bert} Another classic method that uses BERT-based models to generate high-quality sentence embeddings to capture the semantics of sentences. Specifically, we choose roberta-base for initialization.

\textit{\textbf{BGE}}~\citep{bge_embedding} The state-of-the-art method on MTEB\footnote{https://huggingface.co/spaces/mteb/leaderboard}, trained on 300 million text pairs of data, also performed well in the retrieval task. To align the vector dimensions, we chose BGE-base for comparison.

We use RoBERTa-base~\citep{YinhanLiu2019RoBERTaAR} models as the backbone for EER training. The encoder and decoder modules of EBB both come with $12$ Transformer layers (total $24$ layers) and, $768$ hidden size. We explore hyperparameters such as batch size, learning rate, and prompt templates. Finally, We train EER with adam as the optimizer, the batch size as $256$, and the learning rate as $5e^{-5}$. We search for prompt templates, as is discussed in Section~\ref{sec:promptsearch}. For Sentence-BERT and BGE models, we use similar parameters to fine-tune the same data\footnote{We use a popular version of the Chinese Sentence-BERT model available at \url{https://huggingface.co/DMetaSoul/sbert-chinese-general-v2/tree/main}, and the Chinese version of BGE-base available at \url{https://huggingface.co/BAAI/bge-base-zh}.}.

\subsection{Results and Study}
\subsubsection{Overall Results}\label{sec:oresult}

\begin{table}[t]
\centering
\begin{tabular}{c | c c c}
\hline
Models  & R@10  & MRR@10  & AUC  \\
\hline
BM25 & 0.579 & 0.556 & 0.773 \\
Sentence-BERT & 0.693 & 0.650 & 0.827 \\
BGE & 0.771 & 0.694 & 0.915 \\
\hline
EER & \textbf{0.829} & \textbf{0.757} & \textbf{0.931} \\
\hline

\end{tabular}
\caption{Evaluation of EER and baselines.}
\label{tab:exp_result}
\end{table}

Table~\ref{tab:exp_result} shows the comparison results on our disclosed dataset. Our proposed EER achieves better performance than baseline methods. Additionally, we can make the following three observations, which help to understand real-time retrieval and the advantages of EER.

First, as shown in Figure~\ref{fig:demo}, due to the diversity of expressions of the same popular event on the Internet and the simplicity of the query, the solution based on literal matching is not efficient. Compared with BM25, the performance of semantic-based models is significantly ahead. This simultaneously indicates that the dataset is characterized.

Second, EER goes beyond the two semantic-based baselines to demonstrate the excellent performance of the addition of event extraction - the decoder module on the document side.

Third, the performance of BGE surpassed Roberta, which was trained on 300 million data, demonstrating the importance of larger and better quality data. So we believe that disclosing this real-time search data from a real search engine is meaningful for information retrieval research.

\subsubsection{Component Effectiveness Study}

\begin{table}[t]
\centering
\setlength{\tabcolsep}{5pt}
\begin{tabular}{l | c c c}
\hline
Models  & R@10& MRR@10  & AUC  \\
\hline
base & 0.687 & 0.597 & 0.829 \\
base+CL & 0.734 & 0.628 & 0.850 \\
base+CL+GD & 0.769 & 0.673 & 0.884 \\
base+CL+GD+GP & 0.786  & 0.679  & 0.910  \\
base+CL+GD+QER & 0.802 & 0.704 & 0.915 \\
\hline
EER & 0.829 & 0.757 & 0.931 \\
\hline
\end{tabular}
\caption{Evaluation of EER components. CL, GD, GP, and QER are abbreviations for contrastive learning, generative decoder, generative prompt, and relevance learning between query and title, respectively. Base deserves to be Roberta-base.}
\label{tab:ab_exp}
\end{table}

In this section, we discuss further the effectiveness of each component in our model. We make comparisons with the baseline by adding only one component at each time. The results of the experiment are illustrated in Table~\ref{tab:ab_exp}.

\begin{table*}[h]
\centering
\begin{tabular}{l | c c c}
\hline
Templates  & R@10 & MRR@10  & AUC  \\
\hline
In [X], the subject is [MASK] & 0.817  & 0.739  & 0.922  
\\
In [X], the subject is [MASK], the object is [MASK], the action is [MASK] & \textbf{0.829}  & \textbf{0.757} & \textbf{0.931}\\
{[X] $v_1$...[MASK]...$v_n$} & 0.798  & 0.726  & 0.904    \\
{[X] $v_1$...[MASK][MASK][MASK]...$v_n$} & 0.819  & 0.736  & 0.919   \\
\hline
\end{tabular}
\caption{Performance of different prompt templates.}
\label{tab:prompt_search}
\end{table*}

\begin{figure}[t]
    \centering
    \includegraphics[width=\linewidth]{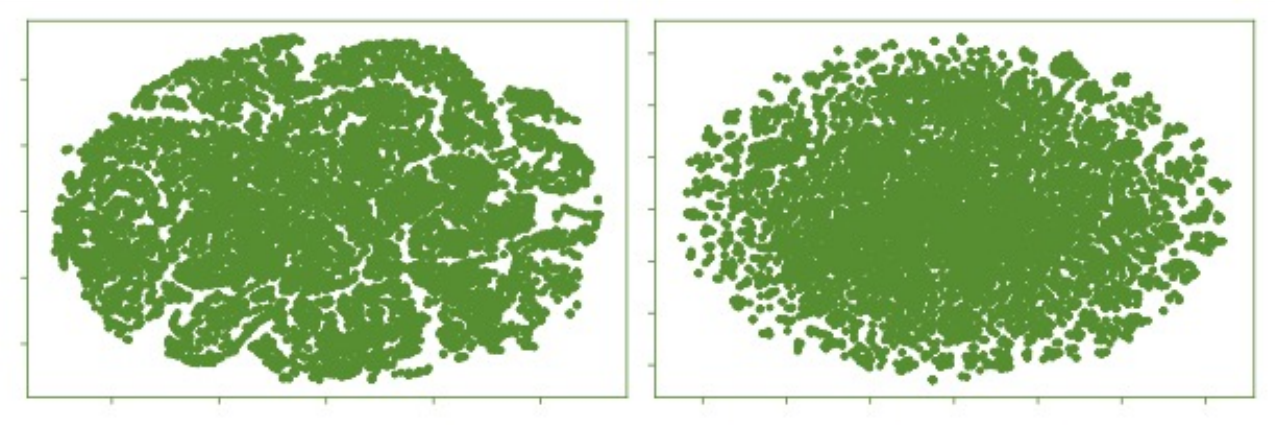}
    \caption{The t-SNE visualization of representations from encoders without and with contrastive learning. As demonstrated in the left part, without contrastive learning, the model encodes queries into a smaller space with more collapses. And on the right, the addition of contrastive learning expands the embedding space with better alignment and uniformity.}
    \label{fig:tsne}
\end{figure}

\begin{figure*}[t]
    \centering
    \includegraphics[width=\linewidth]{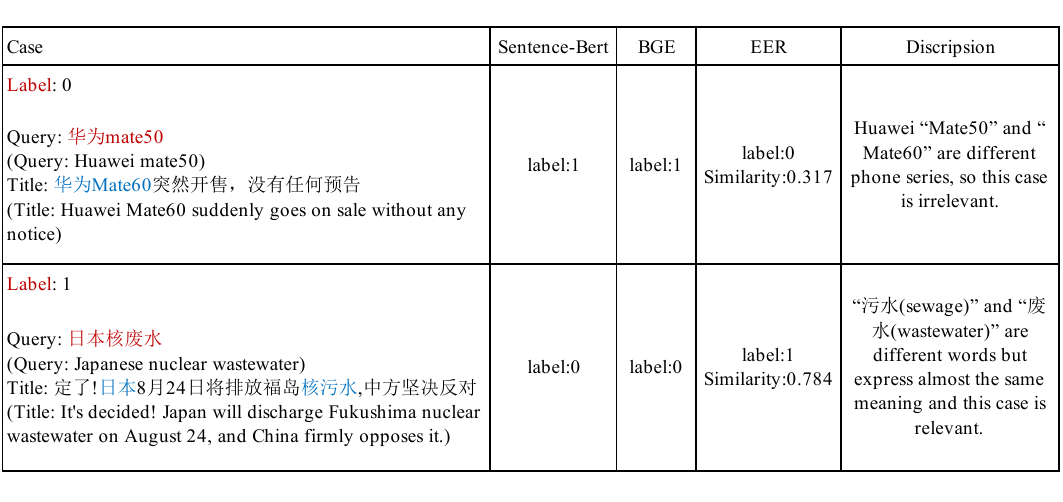}
    \caption{Typical case demonstration of EER and baseline. Relevant query-title pairs are marked as 1 and irrelevant ones as 0.}
    \label{fig:caseh}
\end{figure*}

\textit{\textbf{Contrastive Learning}} Supervised contrastive learning technique is adopted to alleviate representation space degradation problems~\citep{gao2019representation}. To validate its effect, we illustrated the performance in Line $2$~(RoBERTa+CL) of Table~\ref{tab:ab_exp}, where all three metrics, recall, MRR, and AUC, grow significantly compared to the base model. Additionally, to demonstrate that our proposed method can address the issue of representation degradation, we visualize the results of a two-dimensional t-SNE~\citep{van2008visualizing} graph on the embedding of 100,000 queries, which is depicted in Figure~\ref{fig:tsne} and provides further evidence to support our conclusion.

\textit{\textbf{Event Generative}} To make the model implicitly focus on event information, a decoder module is added for event generative learning. As seen in the third line, we learn that decoder~(base+CL+GD) brings $4.8\%$ R@$10$, and $7.1\%$ MRR@$10$ increments on top of the second line. The improvement in metrics in this step is significant.

\textit{\textbf{Prompt Guide}} The fourth line shows that models within the prompt technique get a bit better performance than without versions. Keyword-based prompt learning proves its appeal.

\textit{\textbf{Interaction Between Query and Event}}
The fifth row in Table~\ref{tab:ab_exp} shows that we focused on the impact of adding correlation learning between queries and events~(the results generated by the decoder module) without considering prompt learning~(which would be the full EER model if it were added). The results are obvious - the direct interaction of queries and events effectively contributes to encoder performance, with metrics improving, compared to the third row.

\subsubsection{Prompt Search}\label{sec:promptsearch}

We try to find suitable prompt templates and mainly explore two different types of templates: hand-craft and continuous. Prompt search experimental results are shown in Table~\ref{tab:prompt_search}. Among them, the template ``In \texttt{[X]}, the subject is \texttt{[MASK]}, the object is \texttt{[MASK]}, the action is \texttt{[MASK]}'' performs best. We analyzed that longer and more specific templates strongly imply the key information of a given title, i.e. what the event is, and therefore this hand-craft template is more competitive.

\subsection{Case Study}

To visually illustrate how EER works, we list two typical cases in Figure~\ref{fig:caseh} for qualitative analysis.

In the first case, since the rest of the words in the query are included in the title except for the term “50”, both semantic-based baselines consider the case as relevant, appearing similar to the BM25 algorithm without capturing the huge semantic inconsistency caused by the subtle differences between the terms. In contrast, EER can focus on the fact that the subject of the event in the title is inconsistent with the query and thus makes a distinction.

\begin{figure}[t]
    \centering
    \includegraphics[width=0.5\textwidth,keepaspectratio]{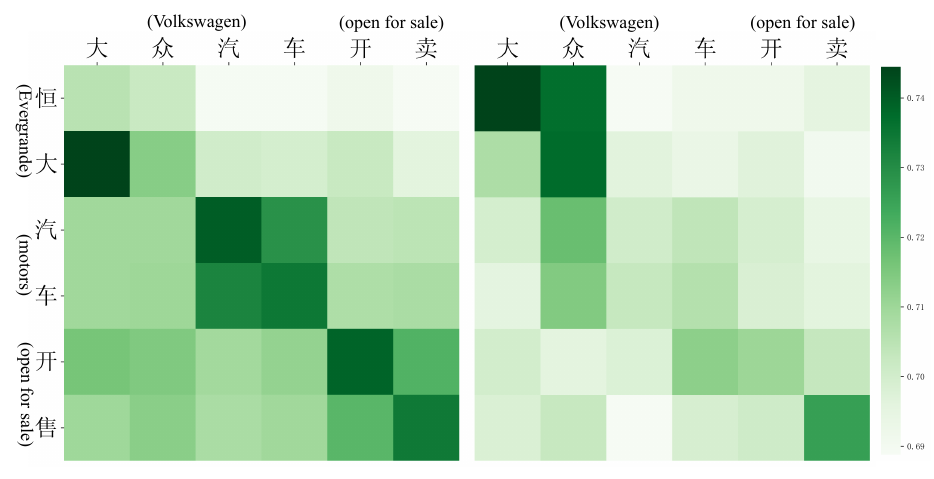}
    \caption{Distribution of Roberta~(a) and our method~(b).}
    \label{fig:hmap}
\end{figure}

In the second case, there are synonym pairs such as ``废水~(sewage)'' and ``污水~(wastewater)'', which the semantic-based model should have taken advantage of. However, the information about the query in the title is very dispersed~(not continuous but scattered), and there is also redundant information ``中方坚决反对~(China firmly opposes it)'' to form interference, which leads to very asymmetric information between the query and the title. Baselines did not make a correct judgment. On the other hand, the event generated by the title is $[$``日本~(Japan)'', ``将排放~(will discharge)'', ``福岛核污水~(Fukushima nuclear sewage)''$]$. Shorter, more focused information facilitates matching to the query, with EER labeled correctly.

\textbf{Attention Distribution} To verify the fusion effect of the decoder module, we plot the attention weight distribution of EER compared to Roberta. As shown in Figure~\ref{fig:hmap}(a), Roberta is more likely to focus on the matching of similar tokens and underestimate the inconsistent parts. In contrast, in Figure~\ref{fig:hmap}(b), with the help of the decoder module, the attention distribution becomes more reasonable, especially the weight between ``恒大(Evergrande)'' and ``大众(Volkswagen)'' increases significantly. This indicates that EER simultaneously emphasizes different parts of the sentence pair.

\section{Related Work}
\subsection{Information Retrieval}
\textit{\textbf{Realtime Search}} Information retrieval~(IR) is a classic NLP task and is widely used in a variety of information-sharing scenarios - after all, people always need to find information. From the perspective of commercial search engine functionality, information retrieval can be categorized into various modes such as knowledge retrieval, product retrieval, code retrieval, and so on~\citep{solvberg1992knowledge,productsearch,codesearch,zhang-etal-2023-event}. Among these modes, some have been extensively and deeply explored by many researchers, while others have not. For example, real-time retrieval has not received much attention. Many events are happening all over the world every moment and are being reported and shared, for example, on Twitter there are hundreds of thousands of tweets every second~\citep{busch2012earlybird}, which makes real-time retrieval very important and used to satisfy the attention of users on new events.

\textit{\textbf{Retrieval Paradigm}} For a long time, researchers have done a lot of exploration. For example, the classical unsupervised approach BM25 mainly focuses on the degree of lexical matching to respond to the match between the query and the document. Neural network models have also been widely used in information retrieval. DSSM~\citep{huang2013learning} uses a deep learning network to map query and document into a semantic space of the same dimension, thus obtaining a low-dimensional semantic vector representation of the utterance sentence embedding, which is used to predict the semantic similarity of two sentences. Poly-encoder~\citep{humeau2019poly} employs multiple independent encoders, each focusing on processing different information, to solve the bi-encoder's low matching quality problem and the slow matching speed of interactive cross-encoders such as ARC-II. The subsequent Colbert~\citep{10.1145/3397271.3401075} structure is relatively streamlined, introducing a late interaction architecture. By delaying and preserving this fine-grained similarity, the ability to pre-compute document representations offline is gained, greatly speeding up queries. \citet{yang2023event} proposes to use event extensions to assist retrieval, requiring additional information to be added.

\textit{\textbf{Sentence Embedding}} 
Improved NLU technology leads to better sentence representation, which is crucial for information retrieval using vector representations. Sentence-BERT~\citep{reimers-2019-sentence-bert}, by using specific fine-tuning techniques, can generate semantically rich sentence embedding representations that achieve better performance in tasks such as text matching. Sentence-T5~\citep{ni2022sentence} adds the decode module for sentence embedding, exploring a variety of representations. \citet{su-etal-2023-one} tries to let the model generate sentence vectors suitable for downstream tasks by giving different instructions to the model, which improves the performance of sentence embedding through more diversified data. In the era of large language models, there have also been some explorations~\citep{jiang2023scaling} of sentence embedding representations with generative models, however, due to the difference between NLU and NLG, related work is still in its infancy. It is important to note that models with large numbers of parameters together with huge amounts of data can cost a lot of money and cause more carbon emissions, so lightweight and low-cost modeling studies are still of great practical relevance.

\subsection{Event Extraction}

Event extraction~\citep{hogenboom2011overview} is the task of organizing natural text into structured events, that is, extracting specific events that occurred at a specific time and place and involved one or more actors, each associated with a set of attributes.

Traditional methods~\citep{ji-grishman-2008-refining, hong-etal-2011-using, li-etal-2013-joint} rely on human-designed features and rules to extract events. The event extraction model based on neural networks~\citep{nguyen-grishman-2015-event,nguyen-etal-2016-joint-event} uses multiple model paradigms for modeling through automatic feature learning. Among them, the most common classification-based method considers event extraction as classifying given trigger and argument candidates into different labels~\citep{feng-etal-2016-language,liu-etal-2018-jointly,lai-etal-2020-event,wang-etal-2021-cleve}. Another sequence tagging method~\citep{chen-etal-2018-collective,ding-etal-2019-event,ma-etal-2020-resource,guzman-nateras-etal-2022-cross} performs EE by tagging each word according to a specific tagging pattern such as BIO~\citep{ramshaw-marcus-1995-text}. With the research on machine reading comprehension tasks, the EE task paradigm has also been transformed into MRC to solve~\citep{wei-etal-2021-trigger,zhou-etal-2022-multi}. This approach employs a span prediction paradigm to predict event triggers and the start and end positions of argument spans. In addition, there are also some works~\citep{huang-etal-2022-multilingual-generative,zeng-etal-2022-ea2e} exploring generating EE result sequences through conditional generation models as well as combining MRC and generative tasks~\citep{deng-etal-2022-title2event}. With the development of huge language models, some studies~\citep{gao2023exploring,wei2023zero} have also discussed the performance of ChatGPT on EE.

\section{Conclusion}

This paper describes an embedding-based approach, called EER, designed to improve semantic retrieval performance in real-time search.
By uniquely utilizing a generative decoder module, our model provides a deeper understanding of the event information implicit in documents, thus enhancing query event matching and significantly reducing the "semantic drift" problem faced in real-time search. We have conducted extensive experiments and analysis to demonstrate the effectiveness of EER. Meanwhile, compared with the currently widely deployed models in real-world scenarios, our model does not bring additional costs because the model parameters are unchanged in the inference stage. Recently LLM has made a big splash in retrieval with its excellent performance, while the high inference cost constrains the wide application of LLM, and of course there are some ongoing works exploring the cost reduction. We believe that our proposed method will bring more thinking perspectives to the field of information retrieval at present.

\nocite{*}
\section{Bibliographical References}\label{sec:reference}

\bibliographystyle{lrec-coling2024-natbib}
\bibliography{lrec-coling2024-example}

\bibliographystylelanguageresource{lrec-coling2024-natbib}
\bibliographylanguageresource{languageresource}

\end{CJK*}
\end{document}